%% file: arxiv_submission_v2.tex
\documentclass{article}

\usepackage{microtype}
\usepackage{graphicx}
\usepackage{subfigure}
\usepackage{booktabs} 

\usepackage{hyperref}

\usepackage[square,comma,authoryear]{natbib}

\newtheorem{thm}{Theorem}

\usepackage{amsmath}
\usepackage{amssymb}
\usepackage{amsfonts,bbm}       
\DeclareMathOperator*{\argmin}{argmin}

\usepackage[table]{xcolor}
\colorlet{mygray}{gray!50!white}

\usepackage[arxiv]{icml2021}

\icmltitlerunning{Representation Transfer by Optimal Transport}

\newlength\figH
\newlength\figW
\usepackage{pgfplots}
\usepackage{xcolor}
\usepackage{multirow}
\usepgfplotslibrary{groupplots}
\usetikzlibrary{pgfplots.groupplots}
\usetikzlibrary{fadings}

\begin{document}
\input{preamble}
\newif\iftwocolumn
\twocolumntrue
\twocolumn[
\icmltitle{Representation Transfer by Optimal Transport}





\begin{icmlauthorlist}
\icmlauthor{Xuhong Li}{utc,baidu}
\icmlauthor{Yves Grandvalet}{utc}
\icmlauthor{Rémi Flamary}{rf}
\icmlauthor{Nicolas Courty}{nc}
\icmlauthor{Dejing Dou}{baidu}
\end{icmlauthorlist}

\icmlaffiliation{utc}{Université de technologie de Compiègne, CNRS, Heudiasyc UMR 7253, France}
\icmlaffiliation{rf}{Université Côte d’Azur, CNRS, OCA, Lagrange UMR 7293, France}
\icmlaffiliation{nc}{Université Bretagne-Sud, CNRS, IRISA UMR 6074, France}
\icmlaffiliation{baidu}{Big Data Laboratory, Baidu Research, Beijing, China}

\icmlcorrespondingauthor{Xuhong Li}{lixuhong@baidu.com}

\icmlkeywords{Machine Learning, ICML}

\vskip 0.3in
]



\printAffiliationsAndNotice{}  

\begin{abstract}
Learning generic representations with deep networks requires massive training samples and significant computer resources.
To learn a new specific task, an important issue is to transfer the generic teacher's representation to a student network.
In this paper, we propose to use a metric between representations that is based on a functional view of neurons.
We use optimal transport to quantify the match between two representations, yielding a distance that embeds some invariances inherent to the representation of deep networks.  
This distance defines a regularizer promoting the similarity of the student's representation with that of the teacher.
Our approach can be used in any learning context where representation transfer is applicable. 
We experiment here on two standard settings: inductive transfer learning, where the  teacher's representation is transferred to a student network of same architecture for a new related task, and knowledge distillation, where the teacher's representation is transferred to a student of simpler architecture for the same task (model compression).
Our approach also lends itself to solving new learning problems; we demonstrate this by showing how to directly transfer the teacher's representation to a simpler architecture student for a new related task.

\end{abstract}

	\section{Introduction}
	
	
	Deep learning requires numerous training samples and significant computing resources for learning good representations of input data; computation resources may also be too limited for evaluation in real-time applications.
	This computational burden, however, can be alleviated by the transfer of representations with transfer learning, or by compressing the representations of large models to smaller ones.
	This type of representation transfer relies on learning protocols that incorporate an implicit or explicit inductive bias towards the initial representation; regularization is a typical scheme for explicitly encouraging such biases. 
	For example, in the situation of fine-tuning from pre-trained weights, the \textit{-SP} parameter regularizers \citep{li2018explicit} constrain the parameters to remain in the vicinity of the initial values, in order to preserve the knowledge encoded in these parameters. 
	
	However, although a neural network is fully described by its parametric function form, regularizing in parameter space is problematic, since the relationship between the parameters of a network and the function implemented by that network does not lend itself easily to analysis; there is no relevant measure of the differences between two deep networks based on their parameters. 
	In this paper, 
	we devise regularizers penalizing the deviations between the internal \textit{representations} of inputs made by two models, one of them being a reference for the trainable one. 

	An internal representation is the encoding from inputs to the features computed at a given layer, that is, the neuron outputs of that layer. 
	Computing relevant similarities between representations thus amounts to comparing these mappings. 
	However, such a comparison should be invariant of the ordering of neurons, which is irrelevant regarding representations.

	We compare representations through \textit{Optimal Transport} (OT).
	OT measures the difference between two sets of elements (neurons in our case), with the crucial property of being invariant to the ordering of these elements, reflecting the property of neural networks to be invariant to permutations of neurons within a layer. 
	We thus propose to transfer representations by encouraging small deviations from a reference representation through an OT-based regularizer.
	
	Our regularization framework suits all the training problems aiming at retaining some knowledge from past or concurrent experiences. 
	It may thus be applied to all the protocols considering training on a series of related distributions, such as domain adaptation or multi-task learning. 
    We validate the effectiveness of the OT-based regularizer in the transfer learning and model compression settings, propose a new setting targeting both transfer and compression in a single step, and provide some analyses on the transport effect.

	\section{A Representation Regularizer}
	A representation is the encoding of input data at a given layer, which results from all the computations performed upstream in the network. 
	The parametric form of this computation being hardly amenable to analyses, we rely on the activations of neurons for comparing representations. 

	\subsection{Rationale}
	
	\paragraph{Distance Between Neurons}
	We view each neuron as a measurable and bounded elementary function whose domain is the network input space and codomain is the output space of this neuron, that is, $a \in L^p$, $a: \mathbb{R}^m \rightarrow \mathbb{R}$. 
	The similarity of two neurons $a$ and $b$ can then be measured by the similarity of their activations, using the $L^p$-norm:
	\begin{align*}
	\|a-b\|_p^p & = \textstyle\int |a(\xvec)-b(\xvec)|^p d\mu_{\xvec} \\
	& = \mathbb{E}_{\xvec \sim {\mu}_{\xvec}} \big[|a(\xvec)-b(\xvec)|^p\big]
	\enspace, 
	\end{align*}
	where ${\mu}_x$, the distribution of the inputs $\xvec$ is used as the measure. 
	In the remainder we choose $p=2$. 
	
	This expectation defines a natural distance between neurons viewed as elementary functions; this distance is compliant with the usual risk that is defined as the expected loss (see Equation~\eqref{eq:generaloptimizationproblem} below). 
	In practice, evaluating the full integral is in general intractable. 
	Instead, the empirical distribution, or a variant thereof, such as the empirical distribution produced by some data augmentation scheme, can be used to compute a proxy for this distance.
	
	\paragraph{Representation as a Neuronal Ensemble}
	\label{subsection:neuron-distribution}
	
	We view neurons as functions, which can be summarized by a vector for practical reasons, but instead of considering representations as the space spanned by these functions (or vectors) at a given layer, we adopt another viewpoint, where a layer is a subset of the neurons of the network.
	The similarity between two subsets of neurons will be assessed by an optimal transport functional, which allows for different subset sizes, is invariant of the permutations of neurons, and is usually robust to outliers and noise \citep{peyre2018computational}.

	\paragraph{Optimal Transport of Neurons}
	In contrast to most uses of optimal transport (OT) in machine learning, the measures defining optimal transport are not interpreted here as probability distributions.
	As a quick reminder, OT considers two discrete measures: $\sum_{i=1}^{n} \mu_i \delta_{\uvec_i}$ and $\sum_{i=1}^m \nu_i \delta_{\vvec_i}$, with $\sum_{i=1}^{n} \mu_i = \sum_{i=1}^{m} \nu_i = 1$, and $\delta_{\uvec}$ is the Dirac measure at $\{\uvec\}$.
	Then, with a defined 
	cost matrix $\bfM \in \Rset^{n \times m}$, 
	Kantorovich’s optimal transport problem reads:
	\begin{equation}
	\label{eq:ot-discret-problem-c4}
	\min_{\bfP \in \Pi(\boldsymbol{\mu}, \boldsymbol{\nu})} \langle\bfP, \bfM\rangle_F
	\enspace,
	\end{equation}
	where $\langle \cdot, \cdot \rangle_{F}$ is the Frobenius product, 
	and $\Pi(\boldsymbol{\mu}, \boldsymbol{\nu}) = \{ \bfP \in \Rset^{n \times m}: \bfP\mathbbm{1}_m=\boldsymbol{\mu} \ \text{and}\ \bfP^T\mathbbm{1}_n=\boldsymbol{\nu}\}$ is the set of all admissible couplings between the two measures ($\mathbbm{1}_m$ is a vector of ones of size $m$).
	We note readily that, when $n=m$, (scaled) permutations are among the admissible couplings.

	The OT problem is a well-known linear program with a computational complexity of $\mathcal{O}(n^3\log(n))$.
	\citet{cuturi2013sinkhorn} promoted 
	a fast approximate solution to the OT problem 
	based on the iterative Sinkhorn-Knopp algorithm, which results in a reduced complexity and can be parallelized.
	%
	The algorithm performs some smoothing that may be detrimental in our setting, {\em e.g.}, regarding permutations.
	So we use a solver relying on a proximal point algorithm IPOT \citep{xie2019fast},  still based on fast Sinkhorn iterations, but converging to the exact solution of Problem \eqref{eq:ot-discret-problem-c4}.

	\paragraph{Structural Risk Minimization}

    Our representation regularizer fits in the structural risk minimization framework by following the general formulation where a regularization functional is added to the empirical risk during learning:
	\begin{equation}\label{eq:generaloptimizationproblem}
	\min_{\weights} \mathbb{E}_{(\xvec,y) \sim \hat{\mu}_{\xvec y}} [\ell(f(\xvec;\weights), y) + \alpha \Omega( \bfT(\xvec), \bfA(\xvec;\weights) ) ] 
	\enspace,
	\end{equation} 
	where 
	$\hat{\mu}_{\xvec y}$ is the empirical joint distribution of data, 
	$f$ is the neural network with parameters $\weights$, 
	$\ell$ is the loss function measuring the discrepancy between the network output and the ground truth $y$, 
	$\Omega$ is the regularizer, 
	and $\alpha$ is the regularization parameter that controls the regularization strength. 
	The regularizer proposed here takes as arguments two representations: a fixed reference representation $\bfT(\xvec)$, obtained from a previously trained model, and the current representation $\bfA(\xvec;\weights)$, which depends on the trainable parameters $\weights$ (more precisely on upstream parameters).
	The arguments of $\bfT$ and $\bfA$ will be omitted in the following for brevity.
	When the learning objective \eqref{eq:generaloptimizationproblem} is optimized by an iterative learning procedure relying on mini-batches, both the loss and the regularizer are computed on the current mini-batch, resulting in a
	Monte Carlo estimate of the empirical risk.

	\subsection{OT Regularizer}
	\label{section:regularizers-on-representations}
	
	The OT regularizer is computed at each iteration of the optimization algorithm, using the $n$ examples of the current mini-batch to characterize neurons. 
	Then, $\bfA^{(t)} \in \Rset^{d \times n}$ denotes the representation of these examples, as provided by the current activations of the $d$ neurons of the penultimate layer at iteration $t$;
	$\bfT \in \Rset^{d' \times n}$ denotes the representation of the same examples on the fixed reference model, as computed by the activations of the $d'$ neurons of its penultimate layer.
	For transfer learning with fine-tuning, $d' = d$ and more specifically $\bfT = \bfA^{(0)}$;  for model compression, we usually have $d<d'$ since the student network is smaller than the teacher network.
	The cost matrix is then defined as:
	\begin{equation}
	\bfM^{(t)}_{ij} = \|\bfA^{(t)}_{i\cdot}-\bfT_{j\cdot}\|_2 \enspace,
	\label{eq:cost-matrix}
	\end{equation}
	where $\bfA^{(t)}_{i\cdot}$ are the activations on the mini-batch of neuron $i$ of the trained model and $\bfT_{j\cdot}$ are the activations of neuron $j$ of the fixed model. 

	Computing the cost on a mini-batch introduces random fluctuations on $\bfM^{(t)}$, which impact the optimal transport plan $\bfP^{(t)}$.
	Both can be considered as stochastic estimations of the ``exact'' cost matrix and transport plan, computed from the whole training sample, at iteration $t$.
	The situation differs from subsampling \citep[see, e.g.,][]{sommerfeld2019optimal}, since here the cost matrix is also approximated.  
	As a result, the expectation of the estimated cost matrix is a smoothed version of the exact matrix that is computed from the entire training sample, resulting in a form of regularization of the exact optimal transport problem. 

	The OT representation regularizer is the optimal transport cost:
	\begin{equation}
	\label{eq:regularizer-ot-classic}
	\Omega_P = \langle\, \bfP^{(t)} ,\bfM^{(t)} \rangle_{F} = \min_{\bfP \in \Pi(\boldsymbol{\mu}, \boldsymbol{\nu})} \langle\bfP, \bfM^{(t)}\rangle_F
	\enspace,
	\end{equation}
	where the optimal transport plan $\bfP^{(t)}$
	is obtained by the IPOT algorithm \citep{xie2019fast}.
	

	\subsection{Learning}

\begin{figure}
\centering
\input{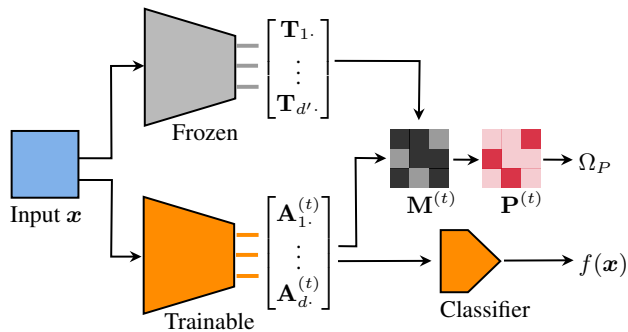}
{\caption{%
			Sketch of representation transfer by optimal transport. 
			The trained network with frozen parameters (gray) and the network with trainable parameters (orange) process the input to compute the reference and trainable representations, $\bfT$ and $\bfA^{(t)}$ respectively.
			Then, the representation regularizer $\Omega_P$ is computed from these representations as in Section \ref{section:regularizers-on-representations}.
}\label{fig:illustration}}
\end{figure}

	The overview of our proposed method is shown in Figure \ref{fig:illustration}.
    During learning, a conceptually simple solution would be to compute the gradients of $\Omega_{P}$ through the Sinkhorn-Knopp iterations.
	For numerical stability and speed, we instead apply the envelope theorem \citep{afriat1971theory,bonnans1998optimization}, which states here that the gradient of $\Omega_{P}$ with respect to $\bfM^{(t)}$ can be calculated at the optimum transport plan $\bfP^{(t)}$, as if $\bfP^{(t)}$ did not depend on $\bfM^{(t)}$.
	
	A caveat is required since, relying on an iterative algorithm, we cannot guarantee that we reached an exact solution $\bfP^{(t)}$.
	In another context, \citet{bach2004multiple} provided guarantees on the approximate optimal solutions reached by similar optimization schemes.
	However, adapting these tools to the problem at hand is still an open issue.
	For completeness, we add that standard computation tricks are required for the numerical stability of gradients through $\bfM^{(t)}_{ij}$ at $\bfM^{(t)}_{ij} = 0$.

    \section{Related Works}

	\paragraph{Representation Transfer}
	Regularizing representations with respect to \textit{data distribution} is very common in general transfer learning settings.
	In domain adaptation, \citet{tzeng2014deep,long2015learning} aim at finding a domain-invariant representations by penalizing the maximum mean discrepancy (MMD) between representations of source and target data.
	\citet{tzeng2015simultaneous} and \citet{li2017learning} propose to record the source knowledge using soft labels of target examples and preserve these soft labels for domain adaptation and lifelong learning respectively.
	\citet{courty2017optimal} formalize domain adaptation as an optimal transport problem by adding a group-sparsity regularizer on the transport plan, and solve it with Sinkhorn algorithm \citep{cuturi2013sinkhorn}.
	More recently, \citet{li2019delta} aligns the feature maps between the pre-trained model and the fine-tuned model using a weighted Euclidean distance, whose weights are obtained by the importance of each neuron, for better transfer knowledge from source to target.
	
	As for model compression and knowledge transfer, where the knowledge accumulated by a large \textit{teacher} model has to be passed on to a small \textit{student} model, knowledge distillation \citet{Hinton2015distilling} aligns the outputs of the teacher and student models, without considering intermediate representations.
	Many works focus on the representation transfer from the teacher model to the student model, {\em e.g.} by directly regressing the student representations from the student model on the teacher's \citep{romero2014fitnets,yim2017gift}, using a variational approximation of the mutual information for pairing the teacher and student representations \citep{ahn2019variational}, or considering the dissimilarities between the representations of the examples of (large) mini-batches, thereby providing invariance with respect to isometries \citep{tung2019similarity,park2019relational}.
	
	\paragraph{Transport of Examples {\em vs} Transport of Features}
	Many works consider samples from the data distribution as in most usages of optimal transport in machine learning \citep{arjovsky2017wasserstein,chen2018adversarial,genevay2018learning,courty2017optimal}.
	We emphasize that our regularizer considers the transportation of features/neurons, which is radically different from the transportation of examples.
	In our framework, the number of neurons $d$ is thus the effective number of samples and the mini-batch size $n$ characterizes the dimension of each neuron. 

	\paragraph{Other Transport of Features}
	Few works have considered transporting neurons. 
	\citet{huang2017like} distilled knowledge to a student model by regularizing feature maps.
	Given one image, they minimized the MMD between two sets of feature maps, in order to transfer the activated regions to the student model.
	This approach differs from ours in that it focuses on the selected/activated regions of a neuron to guide the student model to preserve the region similarities, whereas we focus on knowledge preservation through neuron activations, with similarities that are invariant to permutations to account for neuron swapping.
	
	\citet{singh2019model} consider matching neurons from several trained networks with optimal transport.
	They handle permutations as we do, but their work differs from ours in the following aspects: we are considering a dynamic transfer process where the OT-based regularizer is updated during the transfer process, directly impacting learning, whereas their use of OT mainly consists in finding a static barycenter of two learned models, in order to combine meaningfully those models into a single one, then possibly allowing fine-tuning, but without updating transport in the meantime.
	
	\paragraph{Neurons as Vectors of Activations} 
	We view neurons as functions that are summarized as vectors of activations for practical reasons, and a representation is viewed as a set of neurons.
    This approach of modeling neurons has often been applied, yet mainly for analyzing the representations of neural networks. 
	\citet{li2015convergent} use the statistics of the activations on each neuron to measure the similarity between neurons, and link permuted neurons between independently trained networks with bipartite matching. 
	\citet{raghu2017svcca} propose to combine singular value decomposition and canonical correlation analysis to measure the similarity between the representations of independently trained networks. 
	This modeling is also found in \citet{wang2018towards,morcos2018insights,kornblith2019similarity}. 

	Those works consider a representation as the space spanned by the neuron vectors, whereas we consider discrete sets of neurons: we  assess the dissimilarity between sets of neurons by optimal transport, instead of comparing two spaces spanned by neurons. In this regard, the present view 
	bares some similarities with the one independently proposed by \citet{singh2019model,singh2020model}. Nevertheless, 
	our framework integrates the representation transfer into the learning algorithm, by simply adding the OT regularizer to the empirical risk. Any standard optimization scheme can thus be used to optimize the internal representations of inputs throughout the learning process, jointly with the loss function, whereas \citeauthor{singh2020model} use OT to align the fixed representations of trained networks. The importance of optimizing simultaneously the model coupled with our regularizer is notably elaborated further in Section~\ref{sec:analysis}. 
	In particular, our framework is adapted to the stochastic optimizers used in deep learning.

	\section{Experiments}
	\label{section-experimental-results-ot}
	
	This section provides experimental results assessing the relevance of the OT-based regularizer in three setups.
	We consider two standard learning problems; the first one belongs to the category of transfer learning, where a generic teacher model is used to solve a related task, usually more specific, with a student model. 
	When the student model has the architecture of the teacher model, this task is solved by a fine-tuning strategy, where the teacher model is used to bootstrap the student model.
	The second standard learning problem is model compression, where a complex teacher model is used to train a simpler student model aiming at solving the same task.
	Finally, we show that the OT-based regularizer can also be used for novel applications. 
	Our example here is a transfer learning setup where a complex generic teacher model is used to boost the training of a simpler model for solving a specific task.

	Before presenting these results, we introduce the degraded regularizers that will be used for ablation studies. 
	We recall that $\bfM$ is the cost matrix, $d$ is the size of the reference representation and $d'$ is the size of the learned representation.

	\begin{table*}[t]
		\setlength{\tabcolsep}{10.0pt}
		\centering
		\caption{Average classification accuracy (in \%), using ten-crop test, on transfer learning for 
		no-regularized, $L^2$, $L^2$\textit{-SP}, $\Omega_{I}$, $\Omega_{U}$ 
		and $\Omega_{P}$ regularized fine-tuning. 
		The last column reports the average accuracy on the five tasks.}
		\vspace{2mm}
		\label{table:vision-ten-crop-results}
		\begin{tabular}{@{}lcccccc@{}}
			\toprule
			                           & Aircraft100    & Birds200        & Cars196        & Dogs120        & Foods101    & mean \\ \midrule
			none                       & 83.95$\pm$0.37 & 80.64$\pm$0.30 & 90.21$\pm$0.12 & 69.53$\pm$0.29 & 86.85$\pm$0.09 &  82.18  \\
			$\Omega_{L^2}$             & 83.64$\pm$0.40 & 80.57$\pm$0.36 & 90.51$\pm$0.19 & 69.79$\pm$0.29 & 86.86$\pm$0.09 &  82.27  \\
			$\Omega_{L^2\textit{-SP}}$ & 83.94$\pm$0.39 & 81.10$\pm$0.24 & 90.73$\pm$0.12 & 77.05$\pm$0.19 & \textbf{87.15$\pm$0.14}  & 83.99  \\ 
 			$\Omega_{I}$          & 83.44$\pm$0.45 & 82.25$\pm$0.19 & 90.40$\pm$0.20 & 77.15$\pm$0.17 & 86.88$\pm$0.07 &  84.02  \\
 			$\Omega_{U}$          & 84.74$\pm$0.52 & 81.53$\pm$0.27 & \textbf{91.49$\pm$0.14} & 71.40$\pm$0.28 & 86.90$\pm$0.10 & 83.21  \\
			$\Omega_{P}$ (ours)              & \textbf{85.19$\pm$0.36} & \textbf{82.37$\pm$0.24} & {91.29$\pm$0.13} & \textbf{77.43$\pm$0.13} & {87.06$\pm$0.06} & \textbf{84.67}   \\
			\bottomrule
		\end{tabular}
		\vspace{-2mm}
	\end{table*}
	
	\subsection{Two Degraded Representation Regularizers}
	
    \paragraph{Identity Mapping (No Transport)}
	When $d'=d$, a simple and usual transport plan is to consider that there is no transport, that is, there is no neuron switching between the two representations.
	This amounts to replace the learned transport plan $\bfP^{(t)}$ by the fixed $d\times d$ scaled identity matrix:
	\begin{equation}\label{eq:identitymapping}
		\Omega_I = \big\langle\frac{1}{d} \bfI_{d} ,\bfM^{(t)} \big\rangle_{F}
		\enspace.
	\end{equation}
	This regularizer resembles the $L^2$\textit{-SP} regularizer \citep{li2018explicit}, in the sense that the role of each neuron in the two representations is assumed to be identical.
	Replacing $\Omega_P$ by this regularizer in the experiments shows the value of allowing learned transport plans.

	\paragraph{Uniform Transport}
	We consider another simple fixed transport plan: the uniform joint law between the two sets of neurons, which is the limiting solution for the entropy-regularized OT problem as the regularization parameter goes to infinity \citep[Proposition 4.1]{peyre2018computational}:
	\begin{equation}
		\Omega_U = \big\langle\, \frac{\mathbf{1}_{d\times d'}}{dd'} ,\bfM^{(t)} \big\rangle_{F}
		\enspace,
	\end{equation}
	where $\mathbf{1}_{d\times d'}$ is a $d\times d'$ matrix of ones.
	This regularizer encourages the matching of means, with a variance reduction. 
	Contrary to the identity mapping, it does not assume a specific role to each neuron. It globally encourages all neurons of the learned representation to approach the mean of the reference representation.
	Replacing $\Omega_P$ by this regularizer in the experiments shows the value of looking for a coherent matching between individual neurons, instead of considering the matching of simple summaries of the two populations of neurons.
	
	\subsection{Transfer Learning by Fine-Tuning}
	\label{sec:fine-tuning}

	For fine-tuning, we used ImageNet \citep{deng2009imagenet} as the source task, so that the source representation is sufficiently rich and diverse to be relevant to the target tasks.
	We chose several classification problems on data sets widely used for transfer learning: aircraft models \citep{maji2013fine}, birds \citep{welinderetal2010caltech}, cars \citep{krause20133d}, dogs \citep{khosla2011novel} and food \citep{martinel2016wide}.
	Each target dataset is split into training and test sets following the suggestion of their creators, except for Stanford Dogs 120, whose original test set is a subset of the ImageNet training set. 
	To avoid using the images of the source dataset to evaluate predictions on the target task, we used a part of the ImageNet validation set, which contains only those 120 breeds of dogs, for evaluating the performance on Dogs120. 
	
	We use ResNet \citep{he2016deep} as the backbone network because of its important use in transfer learning, and follow the standard experimental protocol detailed in
	Appendix~A.
	The baselines consist in applying no regularization during fine-tuning, applying the standard $L^2$ norm regularizer (weight-decay) denoted $\Omega_{L^2}$, or penalizing the $L^2$ distance from the pre-trained model $\Omega_{L^2\textit{-SP}}$  \citep{li2018explicit}, where $\textit{-SP}$ refers to the starting point of the optimization process.
	
	There are 33 three-layer residual units in ResNet-101; our OT regularizer is applied to the highest-level representation, that is, the penultimate layer.
	We also penalize an additional layer to preserve intermediate presentations; penalizing the activations of the 19th residual unit performs best among $\{9, 19, 29\}$.
	All regularization hyper-parameters are selected by cross-validation from a range of five logarithmically spaced values from $10^{-4}$ to $1$.

	Table \ref{table:vision-ten-crop-results} shows the results of fine-tuning on the five target datasets.
	We report the average accuracy and its standard deviation on five different runs. 
	Since we use the same data and the same starting point, these runs only differ due to the randomness of stochastic gradient descent and to the random initialization of the parameters from the last layer.
	Our results confirm that the $L^2$\textit{-SP} regularizer is a better choice than the standard weight decay or the absence of regularization.
	Our OT regularizer is always better, except on Foods101, where all methods are within 0.3\%.
	Overall, the benefit of the OT regularizer over $L^2$\textit{-SP} is about half that of $L^2$\textit{-SP} over simple fine-tuning.
	
	The identity mapping $\Omega_{I}$ behaves similarly to $L^2$\textit{-SP}, suggesting that assuming fixed transport plans and asking representations to be alike produces similar effects than constraining the parameters.
	$\Omega_{P}$ is always better than the identity mapping: penalizing the deviations from the reference representation in terms of sets of neurons is beneficial.
	The uniform transport $\Omega_{U}$ is sometimes quite effective, reaching the highest accuracy on Cars196, but sometimes completely unproductive, being very bad on Dogs120.
	On this dataset, the role of the neurons hardly changes during the $L^2\textit{-SP}$ fine-tuning (cf. Figure \ref{figure-sum-diags}), and this is very badly handled by the sole preservation of the global mean of the initial representation.
	In comparison, $\Omega_{P}$ always yields best or close to the best performances: matching individual neurons is a more stable and advantageous strategy than matching populations of neurons.

\subsection{Model Compression}
	

Knowledge distillation aims at compressing a complex model or ensemble into a single smaller model \citep{Hinton2015distilling}.
For this purpose, the training objective follows the general regularized form~\eqref{eq:generaloptimizationproblem}, where the regularization term uses the smoothed response of the large {\em teacher} model as the reference for the smaller {\em student} model: 
%
\begin{align}\label{eq:kd}
  &\Omega_{KD} = KL(g^\tau(\xvec), f^\tau(\xvec;\weights)) \enspace, \\
  &\text{with} \enspace
  g_{k}^\tau(\xvec) = \frac{\exp(h_{k}(\xvec)/\tau)}{\sum_{k=1}^{K}\exp(h_{k}(\xvec)/\tau)}
  \enspace, \nonumber
\end{align}
where $KL(p,q)$ is the Kullback-Leibler divergence of $q$ from $p$, $g^\tau$ and $f^\tau$ are the smoothed responses of the teacher and student model respectively.
The smoothed responses are computed from the vector of activations of the final softmax layer, denoted here $(h_{1}(\xvec),\ldots,h_{K}(\xvec))$ ($K$ is the number of classes).
The so-called temperature $\tau$, which is set to~1 in other contexts, provides smoother responses when set higher, by spreading the non-saturated regions of the softmax, thereby putting stronger emphasis on the boundaries of the classes estimated by the teacher model.
Though several alternatives have been developed, conventional knowledge distillation remains a solid baseline for assessing our representation regularizers \citep[see, e.g.,][]{ruffy2019state,tian2020contrastive}.
We compare here our representation regularizer to knowledge distillation on the CIFAR datasets \citep{krizhevsky2009learning}.
	
	We chose ResNet-1001 \citep{he2016identity} and WRN-28-10 \citep{zagoruyko2016wide} as teacher networks, and  ResNet-56 and a slim version of ResNet-56 as student networks.
	ResNet-56-Slim is identical to ResNet-56 except that it has half as many channels at each layer.
	The temperature used in knowledge distillation is selected from $\{4, 5, 10\}$.
	All regularization hyperparameters are selected from a range of five logarithmically spaced values in $[10^{-3},10]$.
	Other experimental details are deferred to 
	Appendix~A.

	Table \ref{table:cifar} summarizes the results.
\iftwocolumn
		\begin{table}[t]%
		\centering
		\caption{
			Average classification accuracy (in \%) on model compression on CIFAR-10, CIFAR-100 and CIFAR-20 (CIFAR-100 with coarse labels). 
			Letters L, M, S, and W denote respectively the large ResNet-1001, the medium ResNet-56, the small ResNet-56-Slim and the wide WRN-28-10.
			The line labeled ``none'' is for directly training the student model, without teacher.
		}
		\vspace{2mm}
		\label{table:cifar}
			\begin{tabular}{@{}l@{~~~}*{3}{c@{~~~~}}c@{}}
				\toprule
				\multicolumn{1}{@{}c}{CIFAR-10}
				& L $\rightarrow$ S     & L $\rightarrow$ M    & W $\rightarrow$ S     & W $\rightarrow$ M \\ \midrule
				none          & 91.8$\pm$0.2          & 93.6$\pm$0.2         & 91.8$\pm$0.2          & 93.6$\pm$0.2  \\
 				$\Omega_{U}$  & 91.8$\pm$0.2          & 93.8$\pm$0.1         & 91.8$\pm$0.3          & 93.7$\pm$0.2 \\
				$\Omega_{KD}$ & 92.4$\pm$0.1          & 94.0$\pm$0.2         & 92.3$\pm$0.3          & 94.1$\pm$0.1 \\
				$\Omega_P$ (ours)    & \textbf{92.8$\pm$0.1} & \textbf{94.5$\pm$0.1} & \textbf{92.5$\pm$0.2} & \textbf{94.7$\pm$0.1}  \\
				\midrule
				\multicolumn{1}{@{}c}{CIFAR-100}
				& L $\rightarrow$ S     & L $\rightarrow$ M    & W $\rightarrow$ S     & W $\rightarrow$ M \\ \midrule
				none          & 69.6$\pm$0.3          & 73.7$\pm$0.2          & 69.6$\pm$0.3          & 73.7$\pm$0.2          \\
 				$\Omega_{U}$  & 70.0$\pm$0.4          & 74.2$\pm$0.3          & 69.8$\pm$0.3          & 73.9$\pm$0.4          \\
				$\Omega_{KD}$ & \textbf{71.0$\pm$0.1} & 74.8$\pm$0.2          & \textbf{70.7$\pm$0.2} & \textbf{75.6$\pm$0.2} \\
				$\Omega_P$ (ours) & 70.6$\pm$0.2          & \textbf{75.6$\pm$0.1} & 70.5$\pm$0.2          & \textbf{75.6$\pm$0.2} \\
				\midrule
				\multicolumn{1}{@{}c}{CIFAR-20}
				& L $\rightarrow$ S     & L $\rightarrow$ M    & W $\rightarrow$ S     & W $\rightarrow$ M \\ \midrule
				none          &   79.0$\pm$0.2        &   82.3$\pm$0.3       &  79.0$\pm$0.2   & 82.3$\pm$0.3  \\
 				$\Omega_{U}$  & 78.5$\pm$0.2 & 82.1$\pm$0.4  & 78.7$\pm$0.4    & 81.8$\pm$0.5 \\
				$\Omega_{KD}$ & 79.3$\pm$0.5 & 83.2$\pm$0.2 & 79.4$\pm$0.3   & 83.2$\pm$0.4 \\
				$\Omega_P$ (ours)    & \textbf{79.9$\pm$0.1} & \textbf{83.9$\pm$0.2} & \textbf{80.1$\pm$0.4} & \textbf{84.2$\pm$0.1}  \\
				\bottomrule
			\end{tabular}
	\end{table}
\else
	\begin{table*}[t]
		\centering
		\caption{
			Average classification accuracy (in \%) on model compression on CIFAR-10 and CIFAR-100. 
			Letters L, M, S, and W denote respectively the large ResNet-1001, the medium ResNet-56, the small ResNet-56-Slim and the wide WRN-28-10.
			The line labeled ``none'' is for directly training the student model, without teacher.
		}
		\vspace{2mm}
		\label{table:cifar}
			\begin{tabular}{@{}lc@{~~~}c@{~~~}c@{~~~}cc@{~~~}c@{~~~}c@{~~~}c@{}}
				\toprule
				& \multicolumn{4}{c}{CIFAR-10}                                                                 & \multicolumn{4}{c}{CIFAR-100}                                                                  \\
				\cmidrule(lr){2-5} \cmidrule(lr){6-9}

				& L $\rightarrow$ S     & L $\rightarrow$ M    & W $\rightarrow$ S     & W $\rightarrow$ M     & L $\rightarrow$ S     & L $\rightarrow$ M     & W $\rightarrow$ S     & W $\rightarrow$ M     \\ \midrule
				none          & 91.8$\pm$0.2          & 93.6$\pm$0.2         & 91.8$\pm$0.2          & 93.6$\pm$0.2          & 69.6$\pm$0.3          & 73.7$\pm$0.2          & 69.6$\pm$0.3          & 73.7$\pm$0.2          \\
 				$\Omega_{U}$  & 91.8$\pm$0.2          & 93.8$\pm$0.1         & 91.8$\pm$0.3          & 93.7$\pm$0.2          & 70.0$\pm$0.4          & 74.2$\pm$0.3          & 69.8$\pm$0.3          & 73.9$\pm$0.4          \\
				$\Omega_{KD}$ & 92.4$\pm$0.1          & 94.0$\pm$0.2         & 92.3$\pm$0.3          & 94.1$\pm$0.1          & \textbf{71.0$\pm$0.1} & 74.8$\pm$0.2          & \textbf{70.7$\pm$0.2} & \textbf{75.6$\pm$0.2} \\
				$\Omega_P$ (ours)    & \textbf{92.8$\pm$0.1} & \textbf{94.5$\pm$0.1} & \textbf{92.5$\pm$0.2} & \textbf{94.7$\pm$0.1} & 70.6$\pm$0.2          & \textbf{75.6$\pm$0.1} & 70.5$\pm$0.2          & \textbf{75.6$\pm$0.2} \\
				\bottomrule
			\end{tabular}

	\end{table*}
\fi	%
	For CIFAR-10, our representation regularizer consistently improves upon knowledge distillation, with improvements of the order of magnitude of the improvement brought by the knowledge distillation itself upon the student network learned from scratch, without teacher. 
	For CIFAR-100, the results are less decisive: representation transfer tends to perform slightly better for the medium-sized student network whereas knowledge distillation is slightly better for the slim student network. %

	Knowledge distillation examines representations through the lens of smoothed probability estimates. 
	This view is highly relevant for classification purposes, but provides a very approximate summary when the representation space is large relative to the number of classes.
	In these situations, represented here by all the scenarios for CIFAR-10, which has only 10 classes, and, to a lesser extent, by the widest student networks for CIFAR-100 (that is, L\,$\rightarrow$\,M and W\,$\rightarrow$\,M, with learned representation of dimension $d'=256$, to be compared with 100 classes), a more exhaustive transfer of representations is more advantageous. 
	
	CIFAR-20 is the version of CIFAR-100 where 20 coarse labels were defined as superclasses of the 100 "fine" labels by \citet{krizhevsky2009learning}.
	The results on this task support our explanation concerning the link between the relative success of OT on knowledge distillation with the ratio of the dimension of the representation to the number of classes.
	With these superclasses, OT regularization compares more favorably to knowledge distillation,  with improvements that are qualitately similar to CIFAR-10, that is, where OT improves as much on KL as KL improves on learning from scratch. 

	Regarding the degraded variants of representation transfer, 
	we do not report results for the identity mapping $\Omega_I$ that is only applicable when the teacher and student representations are of same size.
%
	The uniform transport $\Omega_{U}$ is much less efficient than  $\Omega_{P}$, with only slight improvements compared to learning the student model from scratch. Here, the mean of the initial representation is too vague an indication to take advantage of the teacher.

	\subsection{Transfer Learning with Model Compression}
	
	Besides the standard transfer learning and model compression setups, the OT-based regularizer can also be used for various new applications. One such application could be model compression in non-classification frameworks, such as regression or ranking, where knowledge distillation needs to be redefined \citep[see, e.g.,][]{takamoto2020efficient,tang2018ranking,chen2017learning}.
	Here we demonstrate this capability in transfer learning from a generic complex teacher model to a simpler student model.
	This transfer is done in a single step, without training the simpler model on the generic task.
    Note that the OT regularizer differs here from the one used for fine-tuning; it is closer to the one used for model compression, in the sense that the reference representation is given by the teacher's model, and not by the student's model.
    
	We chose ResNet-101 trained on ImageNet as the generic and complex teacher network, and MobileNet-V2 \citep{sandler2018mobilenetv2} as the student network. 
	We selected three target classification problems that are neither too close nor too far from ImageNet, that is, Aircraft100, Birds200 and Cars196. 
	The experimental protocol otherwise follows the one of Section \ref{sec:fine-tuning}.
	As there are only a few thousands images in the training sets of the target tasks for 2.2 M parameters, MobileNet-V2 trained from scratch performs badly on all target tasks, and we chose to pre-train MobileNet-V2 on a subset of 10\% of the ImageNet dataset (120\,000  images), for 100 epochs. Using one GPU, this pre-training required 5 hours, which remains frugal regarding computational resources.
	The MobileNet-V2 pre-trained on ImageNet was also fine-tuned for reference purposes.

    Table~\ref{table:mobilenet} displays the main results. 
    The results of pre-training on ImageNet for learning from scratch and for learning with OT regularization are provided here for reference.
    OT regularization has no significant effect when applied on MobileNet with random initialization, and produces small improvements on MobileNet heavily pre-trained on ImageNet (see 
    Appendix~B 
    for full figures).
 \begin{table}[bt]
	\caption{Average classification accuracy (in \%), using 10-crop test, on transfer learning from a complex generic teacher network to a simpler specific student network, where the student network is
	not pre-trained (0\%), pre-trained with 10\% of ImageNet training data, or pre-trained with the ImageNet training data (100\%).}
	\label{table:mobilenet}
	\centering
	\vspace{2mm}
    \begin{tabular}{@{}r@{~~}lccc@{}}
        \toprule
        & & Aircraft100 & Birds200    & Cars196 \\
        \midrule
        0\% &
        none  & {76.23$\pm$0.48} & 57.55$\pm$0.72 & 80.63$\pm$0.39 \\
        \midrule
        \multirow{2}{*}{10\%} &
        none    & 78.49$\pm$0.31 & 71.99$\pm$0.14 & 83.82$\pm$0.15 \\
        & $\Omega_P$ & \textbf{83.48$\pm$0.66} & \textbf{75.17$\pm$0.20} & \textbf{86.36$\pm$0.22} \\
        \midrule
        100\% &
        $\Omega_P$ & {85.82$\pm$0.24} & {80.25$\pm$0.31} & {89.48$\pm$0.23} \\
        \bottomrule
        \end{tabular}
	\vspace{-2mm}
    \end{table}
    The performance of the lightly pre-trained model is about half-way between the random initialization and the heavily pre-trained model.
    In this regime, the OT regularizer boosts performance significantly, by several percent.
    This experiment illustrates that OT regularization enables the transfer of accurate, generic and publicly available representations to ad-hoc networks that would be devised to solve specific classification problems in a reduced amount of time, and with little labeled training data and computing resources.

\section{Analyses of Transport Solutions}
\label{sec:analysis}	
	Regarding run-times, solving the OT problem results in additional computations.
	In our implementation, with mini-batches of 64 examples,  our OT-based regularizer is 2.7  more time consuming than when using a standard parameter regularizer. 
	Using the IPOT algorithm is 1.6 faster than solving directly the linear program for computing the optimal transport Problem \eqref{eq:regularizer-ot-classic}.

	
	\textbf{Impact of Batch Size}. 
	Table \ref{tab:batchsize} displays the impact of batch size on the test accuracy for Aircraft100. 
	For all regularizers, we observe a typical slight detrimental effect of large batch sizes \citep[see, e.g.,][]{keskar2017large}.
	The performance of the OT regularizers is about as stable as the one of the parameter regularizers; their performance decrease is slightly worse for the smallest batch size (16), where the accuracy of all methods decreases, but $\Omega_{P}$ gives the best results for all other sizes. 
	The randomness introduced by small batch sizes in the computation of the transport plans has no visible detrimental effect for batch sizes as small as 32.
	
	\begin{table}[t]
        \caption{Average classification accuracy (in \%), using ten-crop test, on transfer learning on the Aircraft100 target task for 
		$L^2$, $L^2$\textit{-SP}, $\Omega_{I}$, $\Omega_{U}$ 
		and $\Omega_{P}$ regularized fine-tuning.}
		\label{tab:batchsize}
	   \vspace{1mm}
		\scalebox{0.7125}{%
			\begin{tabular}{@{}l@{}lllll@{}}
				\toprule
				\multicolumn{1}{@{}@{}l@{}}{batch size} & \multicolumn{1}{c}{16} & \multicolumn{1}{c}{32} & \multicolumn{1}{c}{64} & \multicolumn{1}{c}{96} & \multicolumn{1}{c@{}}{128} \\ \midrule
				$L^2$        & 79.20$\pm$1.39         & 83.04$\pm$0.23         & 83.64$\pm$0.40         & 83.69$\pm$0.42         & 83.09$\pm$0.41          \\
				$L^2$\textit{-SP}      & \bf 83.40$\pm$0.84         & 83.65$\pm$1.19         & 83.94$\pm$0.39         & 83.04$\pm$0.37         & 82.93$\pm$0.41          \\
				$\Omega_I$          & 82.92$\pm$0.44         & 83.69$\pm$0.57         & 83.44$\pm$0.45         & 82.94$\pm$0.26         & 83.15$\pm$0.41          \\
				$\Omega_U$          & 81.48$\pm$0.33         & 83.95$\pm$0.46         & 84.74$\pm$0.52         & 83.79$\pm$0.33         & 83.36$\pm$0.33          \\
				$\Omega_P$          & 82.31$\pm$0.81         &\bf  84.60$\pm$0.10         & \bf 85.19$\pm$0.36         & \bf 84.60$\pm$0.16         & \bf 83.55$\pm$0.10          \\ \bottomrule
			\end{tabular}
		}
	\end{table}

	\textbf{Is Transport Necessary to Compare Representations?}
	We rely on optimal transport, now used as a tool for post hoc analysis, to quantify the importance of neuron permutations in standard fine-tuning.
	The evolution of the role of neurons before and after fine-tuning is measured by the trace of the transport plan between the representations of the pre-trained model and the model fine-tuned with $L^2$\textit{-SP} regularization that implicitly assumes that neurons maintain their original role throughout learning, and thus penalizes changes in these roles.  
	As OT is used here as an analysis tool, we improve the accuracy of the transport plan by computing the cost matrix on a large representative dataset (3\,000 validation examples).

\setlength\figW{0.41\textwidth}
\setlength\figH{4.1cm}
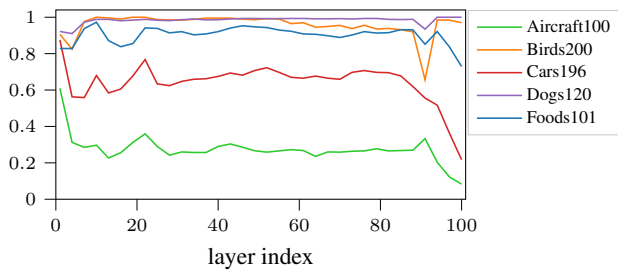
\begin{figure}%
\input{./supple/sum_diags.tex}
	\vspace{-7mm}
{\caption{%
    Ratios of fined-tuned neurons that match their initial activation, as estimated by the traces of the optimal transport plans, versus layer depth. The transport plans are computed, on the validation sets, between the representations formed at each layer of the pre-trained ResNet-101 and those of the model fine-tuned with $L^2$\textit{-SP}.
}\label{figure-sum-diags}}
	\vspace{-2mm}
\end{figure}

	Figure \ref{figure-sum-diags} shows the traces of the transport plans $\mathrm{tr}(\bfP^{(t)})$, obtained by solving Problem \eqref{eq:regularizer-ot-classic}, versus the layer depth for all target tasks considered in Section \ref{sec:fine-tuning}.
	The trace of the transport plan at a given layer is worth one if the activation of each neuron in that layer, after fine-tuning, remains more similar to its initial activation than to the initial activation of any other neuron.
	These traces can thus be interpreted as the ratio of fined-tuned neurons matching their initial activation.
	As expected, the trend is non-increasing with depth: the role of a neuron is more likely to change if the role of its ancestors in the computation graph change.
	Among the five targets, Dogs120 is the only one where these roles remain unchanged throughout the network;
	this is because 
	the classes of Dogs120 belong to ImageNet\footnote{We recall that the test set of Dogs120 was edited to suppress intersections with the training set of ImageNet (see Section  \ref{sec:fine-tuning}).}.
	The dataset Birds200 has also some overlap with ImageNet, but its classification of bird species is more fine-grained;
	the representation of Birds200 after transfer learning is slightly changed.
	As for Foods101, Cars196 and Aircraft100, neurons all along the network have permuted to some extent after transfer learning.
	
	Hence, though $L^2$\textit{-SP} implicitly assumes a perfect matching between the initial and the fine-tuned representation features during fine-tuning, neurons may exchange their roles, possibly in large proportions. 
	This is also likely to apply to all regularization schemes that use the starting point in parameter space as reference \citep[such as][]{li2019delta}, and should be appropriately accounted by the regularizer, as shown by the performances obtained with the identity mapping $\Omega_I$ \eqref{eq:identitymapping}, that are very  similar to the ones of $L^2$\textit{-SP}.  
	
	
	\setlength\figH{5.0cm}
	\setlength\figW{0.475\textwidth}
	\begin{figure}
	\centering
		\input{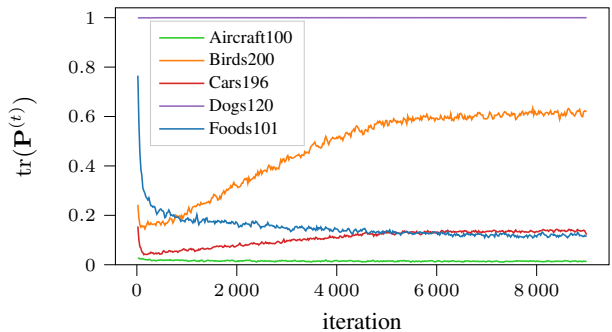}
	    \vspace{-2mm}
		\caption{
		    Ratios of fined-tuned neurons that match their initial activation along training, as estimated by the traces of the optimal transport plans on each mini-batch. 
		    The plans are a by-product of the OT regularizer when fine-tuning with $\Omega_P$.
		}
		\label{figure-diags-training}
	\end{figure}
	
	
	\textbf{Transport Effected}.
	Figure \ref{figure-diags-training} displays the evolution of the same quantity $\mathrm{tr}(\bfP^{(t)})$, on the penultimate layer, during fine-tuning, when explicitly enabling transport with $\Omega_P$.
	As before, the trace of this optimal transport plan is an estimate of the proportion of neuron permutations in that layer, but this estimate is now noisier since it is based only on the few examples in each mini-batch.
	The identity mapping is the void transport plan that leads to the maximal trace of one; except for Dogs120, all computed optimal transport plans differ from the identity: there is an effective transport between neurons.
    The curves vary during training, showing that the alignment between neurons is modified along the optimization, contrary to what would be forced by a fixed matching. Moreover, they vary quite regularly, showing the stability of (traces of) transport plans calculated on mini-batches, and their appropriateness in a stochastic optimization framework.
    
	Regarding tasks, these curves converge towards an ordering similar to that obtained by the $L^2$\textit{-SP} fine-tuning depicted in Figure \ref{figure-sum-diags}, but with lower values that are consistent with the added flexibility of the OT regularizer. 
	Besides, the most striking observation is the sudden drop in similarity that is observed at the beginning of the fine-tuning process. 
	A similar drop is observed on the 19th layer 
	(see Appendix~B),
	except that the values towards which they converge are generally significantly higher, which corresponds to an expectation in transfer learning: generic low-level representations should be even better preserved than the more specific high-level representations.    
	


	\section{Discussion and Conclusion}
	
	A feed-forward neural network learns a series of representations of its inputs at each layer, through the activations of the neurons of that layer.
	Each neuron can be fully described by its analytic parametric expression that is hardly amenable to an analysis: computing a relevant similarity between neurons from their incoming parameters is difficult.
	We thus opt for a more direct functional view of neurons, for which relevant similarities are well-defined.
	
	In previous works, representations are generally formalized as a {\em vector space} of neurons. We rather consider that they are {\em discrete sets} of neurons, to take into account the invariance with respect to the order of neurons in their layer. The Wasserstein distance is therefore a natural candidate to measure of the similarity of representations.
	
	Avoiding factor analysis \citep{raghu2017svcca} to compare representations opens the way to tools that can be directly embedded in stochastic learning algorithms. 
	Our optimal transport regularizer retains the teacher's representation, is invariant to neuron ordering, and tolerant to non-overlapping support of features.
	It works well relative to the state-of-the-art for two existing learning settings, where we show the benefits of transport by comparing it to trivial couplings between the neurons.
    It can operate on representations of different sizes, allowing transfer learning in settings with different teacher and student network architectures, in which fine-tuning is no longer an option. Our framework can also be applied to model compression in regression or ranking frameworks, and we believe that it opens the way to solving many other problems.
	
	
	
	\bibliography{using}
	\bibliographystyle{icml2021}
	
	
	\clearpage
	\appendix

    \renewcommand\thefigure{S\arabic{figure}}
    \setcounter{figure}{0}

    \renewcommand\thetable{S\arabic{table}}
    \setcounter{table}{0}
	
	\section{Experimental Details}
    \label{sec:experimentaldetails}
	We describe here the experimental protocol implemented in the companion source code. 
	
	\subsection{Transfer Learning by Fine-Tuning}

	\subsubsection{Pre-Processing and Post-Processing}
	
	We follow the conventional data processing techniques of \citet{he2016deep}.
	The pre-processing of images involves image resizing and data augmentation.
	We keep the aspect ratio of images and resize the images with the shorter edge being 256.
	We adopt random blur, random mirror and random crop to 224$\times$224 for data augmentation during training.
	Regarding testing, we resize the image in the same ways as training, and then we average the scores of 10 cropped patches (the center patch, the four corner patches, and all their horizontal reflections) for getting the final decision. 
	
	\subsubsection{Optimization}
	
	All optimizations start from a starting point where all layers are initialized with the parameters obtained on ImageNet, except the last layer, whose size is modified according the number of classes in the target task. The latter is initialized randomly.
	
	The optimization solver is Stochastic Gradient Descent (SGD) with momentum 0.9.
	We run for 9000 iterations and divide the learning rate by 10 after 6000 iterations for all target tasks, except for Foods101 for which we run for 16000 iterations and divide the learning rate after 8000 and 12000 iterations.
	The batch size is 64.
	The learning rates are set by cross-validation from $\{0.005, 0.01, 0.02, 0.04\}$.
	
	\subsection{Model Compression}
		
	For both training and testing, we follow the protocol of \citet{he2016deep}.
	For all networks, we run stochastic gradient descent with momentum 0.9, for 160K iterations with batch size 64.
	The learning rate is set to 0.1 at the beginning of training and divided by 10 at 80K and 120K iterations.
	A standard weight decay with a small regularization parameter ($10^{-4}$) is applied for all trainings.
	
    \subsection{Transfer Learning with Model Compression}
    
    The fine-tuning procedure follows the one of Section~\ref{sec:fine-tuning} for both data processing and optimization, except that the reference representation is from a more complex teacher model, and that the student model is either from scratch, lightly pre-trained, or fully pre-trained.

    ResNet-101~\citep{he2016deep} is the teacher model  that was trained on ImageNet, providing the generic representations to be transferred, and MobileNet-V2~\citep{sandler2018mobilenetv2} is the student model to be trained, regularized towards the generic representations of the teacher model through optimal transport.

    The OT-based regularizer significantly improves performance when the student model is lightly pre-trained on ImageNet.
    Specifically, we train MobileNet-V2 with 10\% of the ImageNet training set, with images randomly sampled.
    Training is continued over 100 epochs, with a batch size of 128 and learning rate of 0.1, divided by 10 at the 50th and 75th epochs.
    For interested readers, the lightly pre-trained MobileNet-V2 achieves a top-1 accuracy of 48.76\% on the ImageNet validation set.

	\section{Complementary Experimental Results}
    \label{sec:complementaryresults}
    
    \subsection{Transfer Learning with Model Compression}

    The flexibility of the OT-based regularizer makes it applicable to various setups.
    We have demonstrated this capability in transfer learning from a generic complex teacher to a simpler student model, and here Table~\ref{table:mobilenet-complet} reports our complete figures for transfer learning with model compression, complementing Table~\ref{table:mobilenet}.
    
    The OT-based regularizer $\Omega_P$ significantly improves the performance when the student model is lightly pre-trained, i.e., pre-trained using 10\% ImageNet training data, compared to $\Omega_U$ and no regularization.
    Note that this lightly pre-trained model requires less than 6 GPU-hours and is usable for any transfer learning task.
    
    
    \begin{table}[t]
    	\caption{Average classification accuracy (in \%), using 10-crop test, on transfer learning from a complex generic teacher network to a simpler specific student network, where the student network is
    	not pre-trained (0\%), pre-trained with 10\% of ImageNet training data, or pre-trained with the ImageNet training data (100\%).}
    	\label{table:mobilenet-complet}
    	\centering
    	\vspace{2mm}
        \begin{tabular}{@{}r@{~}llll@{}}
            \toprule
            & & Aircraft100 & Birds200    & Cars196 \\
            \midrule
            \multirow{2}{*}{0\%} &
            none  & \textbf{76.23$\pm$0.48} & 57.55$\pm$0.72 & 80.63$\pm$0.39 \\
            & $\Omega_P$ & {75.79$\pm$0.43} & \textbf{57.90$\pm$0.88} & \textbf{81.26$\pm$0.42} \\
            \midrule
            \multirow{3}{*}{10\%} &
            none    & 78.49$\pm$0.31 & 71.99$\pm$0.14 & 83.82$\pm$0.15 \\
            & $\Omega_U$ & 80.08$\pm$0.48 & 71.88$\pm$0.39 & 85.76$\pm$0.24 \\
            & $\Omega_P$ & \textbf{83.48$\pm$0.66} & \textbf{75.17$\pm$0.20} & \textbf{86.36$\pm$0.22} \\
            \midrule
            \multirow{2}{*}{100\%} &
            none   & 84.96$\pm$0.35 & 79.27$\pm$0.35 & 89.19$\pm$0.10 \\
            & $\Omega_P$ & \textbf{85.82$\pm$0.24} & \textbf{80.25$\pm$0.31} & \textbf{89.48$\pm$0.23} \\
            \bottomrule
            \end{tabular}
    
    \end{table}

    \subsection{Transport Effected}
    
\begin{figure}[t!]
	\centering
		\includegraphics[width=\linewidth]{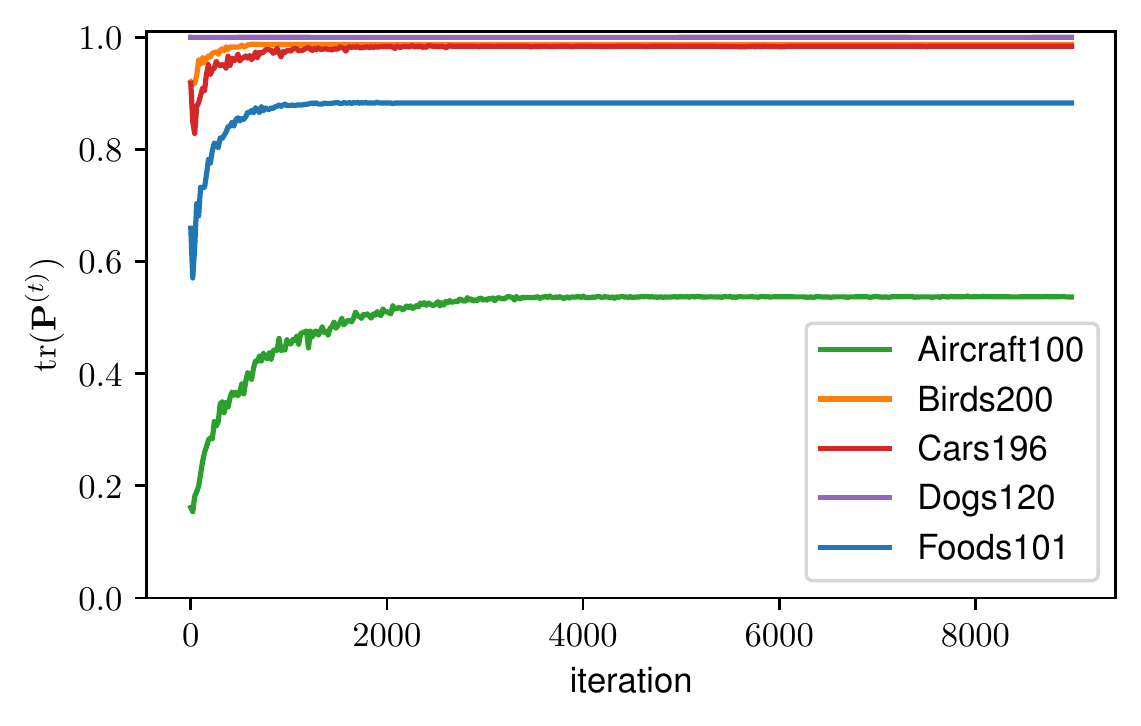}
	    \vspace{-5mm}
		\caption{
		Ratios of fined-tuned neurons at the 19th layer that match their initial activation along training, as estimated by the traces of the optimal transport plans on each mini-batch. 
		}
		\label{figure-diags-training-19}
	\end{figure}

    Figure~\ref{figure-diags-training-19} shows the traces of the transport plans $\mathrm{tr}(\bfP^{(t)})$, on the 19th residual unit, during fine-tuning, when explicitly enabling transport with $\Omega_P$.
    As for the penultimate layer, a sudden drop is also observed at the beginning of the fine-tuning process, but the convergence is much faster and towards higher values.
	
\end{document}

%% file: preamble.tex
\newcommand{\weight}{w}
\newcommand{\weights}{\boldsymbol{\weight}}
\newcommand{\initweight}{\weight^0}
\newcommand{\initweights}{\weights^0}

\newcommand{\funcnet}[1][k]{f_{#1}}
\newcommand{\funcnets}{\boldsymbol{f}}

\newcommand{\group}[1][g]{{\mathcal G}_{#1}}

\newcommand{\source}{\mathcal{S}}
\newcommand{\sourceweight}{\weight}
\newcommand{\sourceweights}{\weights_{\source}}
\newcommand{\newweights}{\weights_{\bar{\source}}}
\newcommand{\sourceinitweights}{\weights^{0}_{\source}}
\newcommand{\sourceinitweight}{\weight^{0}}

\newcommand{\crit}[1][]{J^{#1}}
\newcommand{\loss}{J}

\newcommand{\avec}{\boldsymbol{a}}
\newcommand{\bvec}{\boldsymbol{b}}
\newcommand{\xvec}{\boldsymbol{x}}
\newcommand{\yvec}{\boldsymbol{y}}
\newcommand{\uvec}{\boldsymbol{u}}
\newcommand{\vvec}{\boldsymbol{v}}
\newcommand{\pvec}{\boldsymbol{p}}
\newcommand{\gvec}{\boldsymbol{g}}

\newcommand{\bfM}{\mathbf{M}}
\newcommand{\bfA}{\mathbf{A}}
\newcommand{\bfP}{\mathbf{P}}
\newcommand{\bfF}{\mathbf{F}}
\newcommand{\bfR}{\mathbf{R}}
\newcommand{\bfT}{\mathbf{T}}
\newcommand{\bfS}{\mathbf{S}}
\newcommand{\bfK}{\mathbf{K}}
\newcommand{\bfD}{\mathbf{D}}
\newcommand{\bfI}{\mathbf{I}}
\newcommand{\bfQ}{\mathbf{Q}}

\newcommand{\bfg}{\mathbf{g}}
\newcommand{\bfv}{\mathbf{v}}
\newcommand{\bfG}{\mathbf{G}}
\newcommand{\bfV}{\mathbf{V}}
\newcommand{\bftau}{\boldsymbol{\tau}}
\newcommand{\bfOmega}{\boldsymbol{\Omega}}

\newcommand{\norm}[2][2]{\left\|#2\right\|_{#1}}

\newcommand{\Rset}{\mathbb{R}}
\newcommand{\Esp}{\mathbb{E}}
\newcommand{\Var}{\mathbb{V}}

\newcommand\indep{\perp}

\newcommand{\diag}[1]{{\mathrm{diag}\!\left(\, #1 \right)}}
\newcommand{\transpose}{^\mathsf{T}}
\newcommand{\eqdef}{\stackrel{\triangle}{=}}

\newcommand{\given}[1][]{\:#1\vert\:}

\makeatletter
\def\xinput{%
   \@ifnextchar[%
     {\xinput@i}
     {\xinput@i[j]}%
}
\def\xinput@i[#1]{%
   \@ifnextchar[%
     {\xinput@ii{#1}}
     {\xinput@ii{#1}[i]}%
}
\def\xinput@ii#1[#2]{%
  \@ifempty{#2}%
    {x_{#1}^{\phantom{()}}}
    {\@ifempty{#1}%
    {x_{j}^{(#2)}}
    {x_{#1}^{(#2)}}}%
}
\def\xinputs{%
   \@ifnextchar[%
     {\xinputs@i}
     {\xinputs@i[i]}%
}
\def\xinputs@i[#1]{%
  \@ifempty{#1}%
    {\boldsymbol{x}}
    {\boldsymbol{x}^{(#1)}}%
}
\def\youtput{%
   \@ifnextchar[%
     {\youtput@i}
     {\youtput@i[k]}%
}
\def\youtput@i[#1]{%
   \@ifnextchar[%
     {\youtput@ii{#1}}
     {\youtput@ii{#1}[i]}%
}
\def\youtput@ii#1[#2]{%
  \@ifempty{#2}%
    {y_{#1}^{\phantom{()}}}
    {\@ifempty{#1}%
    {y_{j}^{(#2)}}
    {y_{#1}^{(#2)}}}%
}
\def\youtputs{%
   \@ifnextchar[%
     {\youtputs@i}
     {\youtputs@i[i]}%
}
\def\youtputs@i[#1]{%
  \@ifempty{#1}%
    {\boldsymbol{y}}
    {\boldsymbol{y}^{(#1)}}%
}
\makeatother

%% file: supple/drawing.tex
\footnotesize
\tikzset{every picture/.style={line width=0.75pt}} 

\begin{tikzpicture}[x=0.75pt,y=0.75pt,yscale=-0.56,xscale=0.56]

\draw  [fill={rgb, 255:red, 155; green, 155; blue, 155 }  ,fill opacity=0.69 ] (124,4.07) -- (205.5,28.52) -- (205.5,84.95) -- (124,109.4) -- cycle ;
\draw  [fill={rgb, 255:red, 74; green, 144; blue, 226 }  ,fill opacity=0.7 ] (4.08,114.17) -- (64.25,114.17) -- (64.25,174.34) -- (4.08,174.34) -- cycle ;
\draw  [fill={rgb, 255:red, 255; green, 140; blue, 0 }  ,fill opacity=1 ] (390.5,203.54) -- (415.06,203.54) -- (444.18,231.54) -- (415.06,259.54) -- (390.5,259.54) -- cycle ;
\draw [color={rgb, 255:red, 155; green, 155; blue, 155 }  ,draw opacity=1 ][line width=1.5]    (206.17,55.4) -- (226,55.38) ;
\draw [color={rgb, 255:red, 155; green, 155; blue, 155 }  ,draw opacity=1 ][line width=1.5]    (206.67,74.4) -- (226.5,74.38) ;
\draw [color={rgb, 255:red, 155; green, 155; blue, 155 }  ,draw opacity=1 ][line width=1.5]    (206.67,37.4) -- (226.5,37.38) ;
\draw  [color={rgb, 255:red, 0; green, 0; blue, 0 }  ,draw opacity=1 ][fill={rgb, 255:red, 255; green, 140; blue, 0 }  ,fill opacity=1 ] (123,173.57) -- (204.5,198.02) -- (204.5,254.45) -- (123,278.9) -- cycle ;
\draw [color={rgb, 255:red, 255; green, 140; blue, 0 }  ,draw opacity=1 ][fill={rgb, 255:red, 255; green, 140; blue, 0 }  ,fill opacity=1 ][line width=1.5]    (206.17,225.9) -- (226,225.88) ;
\draw [color={rgb, 255:red, 255; green, 140; blue, 0 }  ,draw opacity=1 ][fill={rgb, 255:red, 255; green, 140; blue, 0 }  ,fill opacity=1 ][line width=1.5]    (206.67,244.9) -- (226.5,244.88) ;
\draw [color={rgb, 255:red, 255; green, 140; blue, 0 }  ,draw opacity=1 ][fill={rgb, 255:red, 255; green, 140; blue, 0 }  ,fill opacity=1 ][line width=1.5]    (206.67,207.9) -- (226.5,207.88) ;
\draw    (298,231.5) -- (379.25,231.26) ;
\draw [shift={(382.25,231.25)}, rotate = 539.8299999999999] [fill={rgb, 255:red, 0; green, 0; blue, 0 }  ][line width=0.08]  [draw opacity=0] (10.72,-5.15) -- (0,0) -- (10.72,5.15) -- (7.12,0) -- cycle    ;
\draw    (64.25,138.75) -- (94,138.8) -- (93.6,55.6) -- (120.6,55.51) ;
\draw [shift={(123.6,55.5)}, rotate = 539.81] [fill={rgb, 255:red, 0; green, 0; blue, 0 }  ][line width=0.08]  [draw opacity=0] (10.72,-5.15) -- (0,0) -- (10.72,5.15) -- (7.12,0) -- cycle    ;
\draw    (64.65,158.35) -- (94.4,158.4) -- (93.73,227.43) -- (118.07,227.43) ;
\draw [shift={(121.07,227.43)}, rotate = 180] [fill={rgb, 255:red, 0; green, 0; blue, 0 }  ][line width=0.08]  [draw opacity=0] (10.72,-5.15) -- (0,0) -- (10.72,5.15) -- (7.12,0) -- cycle    ;
\draw    (294.98,50.98) -- (371.5,51.17) -- (370.87,101.5) ;
\draw [shift={(370.83,104.5)}, rotate = 270.72] [fill={rgb, 255:red, 0; green, 0; blue, 0 }  ][line width=0.08]  [draw opacity=0] (10.72,-5.15) -- (0,0) -- (10.72,5.15) -- (7.12,0) -- cycle    ;
\draw  [color={rgb, 255:red, 255; green, 255; blue, 255 }  ,draw opacity=0 ][fill={rgb, 255:red, 0; green, 0; blue, 0 }  ,fill opacity=0.8 ] (344.6,113) -- (362.2,113) -- (362.2,130.6) -- (344.6,130.6) -- cycle ;
\draw  [color={rgb, 255:red, 255; green, 255; blue, 255 }  ,draw opacity=0 ][fill={rgb, 255:red, 155; green, 155; blue, 155 }  ,fill opacity=1 ] (379.8,113) -- (397.4,113) -- (397.4,130.6) -- (379.8,130.6) -- cycle ;
\draw  [color={rgb, 255:red, 255; green, 255; blue, 255 }  ,draw opacity=0 ][fill={rgb, 255:red, 0; green, 0; blue, 0 }  ,fill opacity=0.8 ] (362.2,113) -- (379.8,113) -- (379.8,130.6) -- (362.2,130.6) -- cycle ;
\draw  [color={rgb, 255:red, 255; green, 255; blue, 255 }  ,draw opacity=0 ][fill={rgb, 255:red, 155; green, 155; blue, 155 }  ,fill opacity=1 ] (344.62,130.58) -- (362.22,130.58) -- (362.22,148.18) -- (344.62,148.18) -- cycle ;
\draw  [color={rgb, 255:red, 255; green, 255; blue, 255 }  ,draw opacity=0 ][fill={rgb, 255:red, 0; green, 0; blue, 0 }  ,fill opacity=0.8 ] (379.82,130.58) -- (397.42,130.58) -- (397.42,148.18) -- (379.82,148.18) -- cycle ;
\draw  [color={rgb, 255:red, 255; green, 255; blue, 255 }  ,draw opacity=0 ][fill={rgb, 255:red, 0; green, 0; blue, 0 }  ,fill opacity=0.8 ] (362.22,130.58) -- (379.82,130.58) -- (379.82,148.18) -- (362.22,148.18) -- cycle ;
\draw  [color={rgb, 255:red, 255; green, 255; blue, 255 }  ,draw opacity=0 ][fill={rgb, 255:red, 0; green, 0; blue, 0 }  ,fill opacity=0.8 ] (344.62,148.18) -- (362.22,148.18) -- (362.22,165.78) -- (344.62,165.78) -- cycle ;
\draw  [color={rgb, 255:red, 255; green, 255; blue, 255 }  ,draw opacity=0 ][fill={rgb, 255:red, 0; green, 0; blue, 0 }  ,fill opacity=0.8 ] (379.82,148.18) -- (397.42,148.18) -- (397.42,165.78) -- (379.82,165.78) -- cycle ;
\draw  [color={rgb, 255:red, 255; green, 255; blue, 255 }  ,draw opacity=0 ][fill={rgb, 255:red, 155; green, 155; blue, 155 }  ,fill opacity=1 ] (362.22,148.18) -- (379.82,148.18) -- (379.82,165.78) -- (362.22,165.78) -- cycle ;
\draw  [color={rgb, 255:red, 255; green, 255; blue, 255 }  ,draw opacity=0 ][fill={rgb, 255:red, 208; green, 2; blue, 27 }  ,fill opacity=0.2 ] (427.65,112.87) -- (445.25,112.87) -- (445.25,130.47) -- (427.65,130.47) -- cycle ;
\draw  [color={rgb, 255:red, 255; green, 255; blue, 255 }  ,draw opacity=0 ][fill={rgb, 255:red, 208; green, 2; blue, 27 }  ,fill opacity=0.8 ] (462.85,112.87) -- (480.45,112.87) -- (480.45,130.47) -- (462.85,130.47) -- cycle ;
\draw  [color={rgb, 255:red, 255; green, 255; blue, 255 }  ,draw opacity=0 ][fill={rgb, 255:red, 208; green, 2; blue, 27 }  ,fill opacity=0.2 ] (445.25,112.87) -- (462.85,112.87) -- (462.85,130.47) -- (445.25,130.47) -- cycle ;
\draw  [color={rgb, 255:red, 255; green, 255; blue, 255 }  ,draw opacity=0 ][fill={rgb, 255:red, 208; green, 2; blue, 27 }  ,fill opacity=0.8 ] (427.65,130.47) -- (445.25,130.47) -- (445.25,148.07) -- (427.65,148.07) -- cycle ;
\draw  [color={rgb, 255:red, 255; green, 255; blue, 255 }  ,draw opacity=0 ][fill={rgb, 255:red, 208; green, 2; blue, 27 }  ,fill opacity=0.2 ] (462.87,130.44) -- (480.47,130.44) -- (480.47,148.04) -- (462.87,148.04) -- cycle ;
\draw  [color={rgb, 255:red, 255; green, 255; blue, 255 }  ,draw opacity=0 ][fill={rgb, 255:red, 208; green, 2; blue, 27 }  ,fill opacity=0.2 ] (445.27,130.44) -- (462.87,130.44) -- (462.87,148.04) -- (445.27,148.04) -- cycle ;
\draw  [color={rgb, 255:red, 255; green, 255; blue, 255 }  ,draw opacity=0 ][fill={rgb, 255:red, 208; green, 2; blue, 27 }  ,fill opacity=0.2 ] (427.67,148.04) -- (445.27,148.04) -- (445.27,165.64) -- (427.67,165.64) -- cycle ;
\draw  [color={rgb, 255:red, 255; green, 255; blue, 255 }  ,draw opacity=0 ][fill={rgb, 255:red, 208; green, 2; blue, 27 }  ,fill opacity=0.2 ] (462.87,148.04) -- (480.47,148.04) -- (480.47,165.64) -- (462.87,165.64) -- cycle ;
\draw  [color={rgb, 255:red, 255; green, 255; blue, 255 }  ,draw opacity=0 ][fill={rgb, 255:red, 208; green, 2; blue, 27 }  ,fill opacity=0.8 ] (445.27,148.04) -- (462.87,148.04) -- (462.87,165.64) -- (445.27,165.64) -- cycle ;
\draw    (298.32,217.98) -- (314.5,218.17) -- (313.83,138.5) -- (338.5,138.8) ;
\draw [shift={(341.5,138.83)}, rotate = 180.69] [fill={rgb, 255:red, 0; green, 0; blue, 0 }  ][line width=0.08]  [draw opacity=0] (10.72,-5.15) -- (0,0) -- (10.72,5.15) -- (7.12,0) -- cycle    ;
\draw    (447.5,231) -- (507.25,230.76) ;
\draw [shift={(510.25,230.75)}, rotate = 539.77] [fill={rgb, 255:red, 0; green, 0; blue, 0 }  ][line width=0.08]  [draw opacity=0] (10.72,-5.15) -- (0,0) -- (10.72,5.15) -- (7.12,0) -- cycle    ;
\draw    (401.83,140) -- (422.17,140.15) ;
\draw [shift={(425.17,140.17)}, rotate = 180.41] [fill={rgb, 255:red, 0; green, 0; blue, 0 }  ][line width=0.08]  [draw opacity=0] (10.72,-5.15) -- (0,0) -- (10.72,5.15) -- (7.12,0) -- cycle    ;
\draw    (482.83,140) -- (506.5,140.15) ;
\draw [shift={(509.5,140.17)}, rotate = 180.36] [fill={rgb, 255:red, 0; green, 0; blue, 0 }  ][line width=0.08]  [draw opacity=0] (10.72,-5.15) -- (0,0) -- (10.72,5.15) -- (7.12,0) -- cycle    ;

\draw (512,222) node [anchor=north west][inner sep=0.75pt]   [align=left] {$\displaystyle f(\xvec) $};
\draw (225,10) node [anchor=north west][inner sep=0.75pt]    {$\mathit{\begin{bmatrix}
		\mathbf{T}_{1\cdot } \\
		\vdots \\
		\mathbf{T}_{d' \cdot }
		\end{bmatrix}}$};
\draw (225,170) node [anchor=north west][inner sep=0.75pt]    {$\begin{bmatrix}
	\mathbf{A}^{( t)}_{1\cdot } \ \\
	\vdots \\
	\mathbf{A}^{( t)}_{d \cdot }
	\end{bmatrix}$};
\draw (0,179) node [anchor=north west][inner sep=0.75pt]   [align=left] {{Input $\xvec$}};
\draw (146,105) node [anchor=north west][inner sep=0.75pt]   [align=left] {{Frozen}};
\draw (139,275.33) node [anchor=north west][inner sep=0.75pt]   [align=left] {{Trainable}};
\draw (387.33,265) node [anchor=north west][inner sep=0.75pt]   [align=left] {{Classifier}};
\draw (357,165.7) node [anchor=north west][inner sep=0.75pt]    {$\mathbf{M}^{( t)}$};
\draw (441.67,165.7) node [anchor=north west][inner sep=0.75pt]    {$\mathbf{P}^{( t)}$};
\draw (512,132) node [anchor=north west][inner sep=0.75pt]    {$\Omega_{P}$};

\end{tikzpicture}

%% file: supple/sum_diags.tex
\begin{tikzpicture}

\definecolor{color0}{rgb}{0.12156862745098,0.466666666666667,0.705882352941177}
\definecolor{color1}{rgb}{1,0.498039215686275,0.0549019607843137}
\definecolor{color2}{rgb}{0.172549019607843,0.8,0.172549019607843}
\definecolor{color3}{rgb}{0.83921568627451,0.152941176470588,0.156862745098039}
\definecolor{color4}{rgb}{0.580392156862745,0.403921568627451,0.741176470588235}
\footnotesize
\begin{axis}[
height=\figH,
legend cell align={left},
legend style={at={(1.005,1.0)}, anchor=north west, draw=white!80.0!black, nodes={scale=0.75, transform shape},font=\footnotesize},
tick align=outside,
tick pos=left,
width=\figW,
x grid style={white!69.01960784313725!black},
xmin=0, xmax=101,
xtick style={color=black},
xlabel={layer index},
label style={font=\small},
tick label style={font=\scriptsize},
y grid style={white!69.01960784313725!black},
ymin=0.037158203125, ymax=1.045849609375,
ymin=0.0, ymax=1.04,
ytick style={color=black}
]
\addplot [semithick, color2]
table {%
1 0.609375
4 0.3125
7 0.28515625
10 0.296875
13 0.2265625
16 0.255859375
19 0.3125
22 0.359375
25 0.2900390625
28 0.2421875
31 0.259765625
34 0.2568359375
37 0.2568359375
40 0.2900390625
43 0.3037109375
46 0.2861328125
49 0.2666015625
52 0.2587890625
55 0.265625
58 0.2724609375
61 0.267578125
64 0.2353515625
67 0.259765625
70 0.2587890625
73 0.263671875
76 0.265625
79 0.27734375
82 0.265625
85 0.267578125
88 0.26953125
91 0.3330078125
94 0.203125
97 0.12255859375
100 0.0830078125
};
\addlegendentry{Aircraft100}
\addplot [semithick, color1]
table {%
1 0.90625
4 0.82421875
7 0.9765625
10 1
13 0.99609375
16 0.990234375
19 1
22 1
25 0.9873046875
28 0.9853515625
31 0.984375
34 0.98828125
37 0.9951171875
40 0.9951171875
43 0.9951171875
46 0.9892578125
49 0.986328125
52 0.9921875
55 0.990234375
58 0.96484375
61 0.9697265625
64 0.9453125
67 0.94921875
70 0.955078125
73 0.9375
76 0.9560546875
79 0.935546875
82 0.9384765625
85 0.931640625
88 0.919921875
91 0.65625
94 0.98486328125
97 0.98388671875
100 0.97021484375
};
\addlegendentry{Birds200}
\addplot [semithick, color3]
table {%
1 0.875
4 0.5625
7 0.55859375
10 0.6796875
13 0.583984375
16 0.60546875
19 0.677734375
22 0.767578125
25 0.6337890625
28 0.6240234375
31 0.6474609375
34 0.6591796875
37 0.662109375
40 0.6748046875
43 0.693359375
46 0.681640625
49 0.70703125
52 0.72265625
55 0.697265625
58 0.669921875
61 0.6650390625
64 0.6767578125
67 0.6650390625
70 0.6591796875
73 0.697265625
76 0.70703125
79 0.697265625
82 0.6953125
85 0.677734375
88 0.619140625
91 0.5556640625
94 0.51708984375
97 0.36328125
100 0.21728515625
};
\addlegendentry{Cars196}
\addplot [semithick, color4]
table {%
1 0.921875
4 0.91015625
7 0.97265625
10 0.98828125
13 0.98828125
16 0.98046875
19 0.984375
22 0.98828125
25 0.9833984375
28 0.9814453125
31 0.986328125
34 0.990234375
37 0.9853515625
40 0.986328125
43 0.990234375
46 0.9931640625
49 0.9931640625
52 0.9921875
55 0.9921875
58 0.9931640625
61 0.9931640625
64 0.9912109375
67 0.9912109375
70 0.9921875
73 0.990234375
76 0.9931640625
79 0.9931640625
82 0.98828125
85 0.9873046875
88 0.98828125
91 0.9345703125
94 1
97 1
100 1
};
\addlegendentry{Dogs120}
\addplot [semithick, color0]
table {%
1 0.828125
4 0.828125
7 0.9375
10 0.97265625
13 0.87109375
16 0.837890625
19 0.85546875
22 0.94140625
25 0.9384765625
28 0.9140625
31 0.9208984375
34 0.9033203125
37 0.908203125
40 0.9208984375
43 0.9404296875
46 0.953125
49 0.947265625
52 0.943359375
55 0.9296875
58 0.9228515625
61 0.908203125
64 0.90625
67 0.8984375
70 0.888671875
73 0.90234375
76 0.9208984375
79 0.9130859375
82 0.9150390625
85 0.931640625
88 0.9306640625
91 0.8515625
94 0.92138671875
97 0.8369140625
100 0.72900390625
};
\addlegendentry{Foods101}
\end{axis}

\end{tikzpicture}